\pdfoutput=1
\documentclass[11pt]{article}

\usepackage[]{ACL2023}

\usepackage{times}
\usepackage{latexsym}

\usepackage[T1]{fontenc}
\usepackage[utf8]{inputenc}

\usepackage{microtype}

\usepackage{inconsolata}

\usepackage{graphicx} %images
\usepackage{multirow}

\title{Adaptive Question Answering: Enhancing Language Model Proficiency for Addressing Knowledge Conflicts with Source Citations}

\author{Sagi Shaier,$^\nabla$ Ari Kobren,$^\dag$ Philip V. Ogren$^\dag$ \\
  $^\nabla$University of Colorado Boulder\\
$^\dag$Oracle \\
$^\nabla$E-mail: sagi.shaier@colorado.edu \\
$^\dag$E-mail: \{ari.kobren, philip.ogren\}@oracle.com \\
 \\}

\begin{document}
\maketitle
\begin{abstract}
Resolving knowledge conflicts is a crucial challenge in Question Answering (QA) tasks, as the internet contains numerous conflicting facts and opinions. While some research has made progress in tackling ambiguous settings where multiple valid answers exist, these approaches often neglect to provide source citations, leaving users to evaluate the factuality of each answer. On the other hand, existing work on citation generation has focused on unambiguous settings with single answers, failing to address the complexity of real-world scenarios. Despite the importance of both aspects, no prior research has combined them, leaving a significant gap in the development of QA systems. In this work, we bridge this gap by proposing the novel task of QA with source citation in ambiguous settings, where multiple valid answers exist. To facilitate research in this area, we create a comprehensive framework consisting of: (1) five novel datasets, obtained by augmenting three existing reading comprehension datasets with citation meta-data across various ambiguous settings, such as distractors and paraphrasing; (2) the first  ambiguous multi-hop QA dataset featuring real-world, naturally occurring contexts; (3) two new metrics to evaluate models’ performances; and (4) several strong baselines using rule-based, prompting, and finetuning approaches over five large language models. We hope that this new task, datasets, metrics, and baselines will inspire the community to push the boundaries of QA research and develop more trustworthy and interpretable systems. Code and data can be found here: \url{https://github.com/Shaier/Adaptive_QA.git}.
\end{abstract}

\section{Introduction}
Knowledge-enhanced large language models (LLMs) have demonstrated remarkable question-answering (QA) capabilities, partially due to their ability to reason over a substantial number of tokens \cite{hu2023survey, wei2021knowledge}. While some work has shown that LLMs do not fully utilize long sequences \cite{liu2023lost}, an issue that arises \textit{from the context itself} is that knowledge is dynamic and is constantly changing, and hence, conflicting facts and opinions may exist within it \cite{min2020ambigqa, neeman-etal-2023-disentqa}. For example, since politicians change, there exist documents each expressing that a different person is the current president of the United states.

This is especially problematic for models that can handle long contexts, such as existing state-of-the-art models \cite{longchat2023, openai2023, mpt_2023, openai2023, anthropic} and retrieval-augmented generation (RAG) systems \cite{rag, wei2021knowledge, hu2023survey}, as the greater the length of the context, the higher the probability of encountering conflicting information from various sources, domains, or even time periods within the same source or domain.

\begin{figure}[t]
\includegraphics[width=1\columnwidth,height=0.7\columnwidth,keepaspectratio]{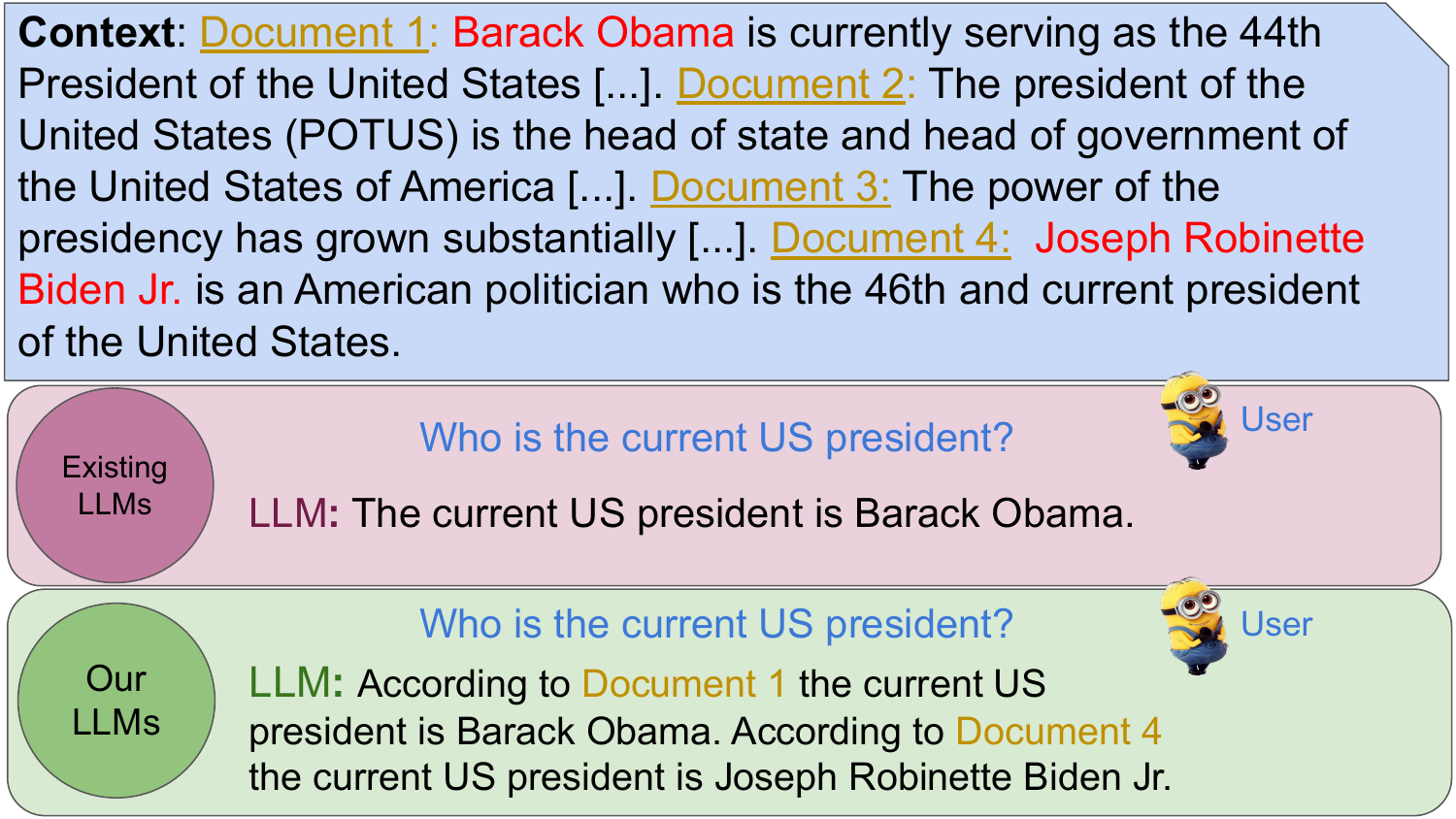}
    \caption{
    % While existing models can answer many questions correctly, when provided with conflicting facts we find that they often choose just one of the facts as the answer. This in turn hinders users from knowing that the context contains multiple answers. In comparison, our approach results in LLMs that produce a distinct response for each conflicting fact \textbf{and also} cite their sources by directing users to where the information came from, which allows for factuality verification and leads to an increase in users' trust and interpretability.
    When faced with ambiguous settings, unlike existing models that often provide a single answer, our methods generate multiple answers and cite their sources, allowing users to verify the answers' factuality and make informed decisions.
    }
    \label{main_figure}
\end{figure}

% conflicting setting
Existing work on QA in contextual\footnote{Unlike settings where context contradicts model knowledge \cite{neeman-etal-2023-disentqa}, which is not our focus.} knowledge conflicts\footnote{Knowledge conflicts occur when multiple answers are possible from a set of documents and a question.} setting mitigate this issue by either predicting all valid answers \cite{min-etal-2020-ambigqa, min-etal-2021-joint}, aggregating all answers \cite{shao-huang-2022-answering, gao-etal-2021-answering}, asking clarification questions \cite{zhang2023clarify}, and other methods \cite{cole2023selectively, sun-etal-2023-answering}. However, these methods burden users with the task of extensively evaluating the factuality of each answer \cite{rawte-etal-2023-troubling, shaier-etal-2023-stochastic, dziri-etal-2022-origin}, \textbf{as they do not cite the source of the answer.}

Hence, it is crucial to develop systems that not only generate all possible answers, but produce a distinct response for each conflicting information while also citing their sources, as shown in Figure \ref{main_figure}. This, in turn, will also lead to an increase in users' trust and interpretability \cite{shaier-etal-2023-stochastic}. And while some existing work develop models that cite their sources, \textbf{they only focus on unambiguous setting}, where only one answer exists \cite{bohnet2023attributed, gao-etal-2023-rarr, slobodkin2024attribute}. Furthermore, \textbf{none of the existing work on contextual conflicts or citation generation focus on complex QA settings}, which require multi-hop reasoning and many answers, and resemble a more realistic real-world setting \cite{joshi-etal-2017-triviaqa, mohammadi2022comprehensive, pan-etal-2023-qacheck}

We bridge the gap between ambiguous QA and citation generation by proposing the novel task of QA with source citation in ambiguous settings, where multiple valid answers exist. To facilitate research, we provide a comprehensive framework featuring: five novel datasets with citation metadata, the first ambiguous multi-hop QA dataset, two new evaluation metrics, and strong baselines. Our goal is to inspire the community to push the boundaries of QA research and develop more trustworthy and interpretable systems.

\section{Related Work}
\paragraph{Knowledge-enhanced Language Models}
% Many recent works have shown that enhancing LMs with external knowledge, such as text documents, allows them to outperform many state-of-the-art baselines. Such systems include retrieval-augmented systems \cite{lewis2021retrievalaugmented, jiang-etal-2023-active, shao-etal-2023-enhancing} and also knowledge-enhanced systems \cite{lymperaiou2024survey, arora-etal-2022-metadata, wang-etal-2020-knowledge, liu2024augment, du2022knowledgeenhanced, deng-etal-2023-knowledge, hu2023survey, wei2021knowledge}. Such knowledge can be provided to models in various settings, such as in reading comprehension, and also from users or retrieval systems. While early variations of these systems had a limited context size, existing work has increased that significantly \cite{longchat2023, openai2023, mpt_2023, openai2023gpt4, anthropic}, which allows them to utilize an immense amount of external knowledge. However, the greater the length of the context, the higher the probability of encountering conflicting information.

Recent works have shown that incorporating external knowledge into LLMs improves their performance. Examples include RAG \cite{lewis2021retrievalaugmented, jiang-etal-2023-active, shao-etal-2023-enhancing} and knowledge-enhanced systems 
% \cite{lymperaiou2024survey, arora-etal-2022-metadata, wang-etal-2020-knowledge, liu2024augment, deng-etal-2023-knowledge, 
\cite{du2022knowledgeenhanced, hu2023survey, wei2021knowledge, shaier2022mindknowledgegapsurvey}. 
% This external knowledge can come from various sources and settings, such as reading comprehension or user input.
While initial systems had limited context, recent advancements have significantly increased context size, allowing for more extensive knowledge utilization \cite{longchat2023, openai2023, mpt_2023, openai2023gpt4, anthropic}. However, this increase also raises the likelihood of encountering conflicting information.

\paragraph{Knowledge Conflicts}
% The greater the length of the context, the higher the probability of encountering conflicting information from various sources, domains, or even time periods within the same source or domain. To that end, 
% Several works have focused on QA in knowledge conflicts setting, which can be split into two categories: the first, focus on ambiguous questions, where depending on how the questions are being asked the answers may vary \cite{min2020ambigqa, sun-etal-2023-answering, cole2023selectively}. The second, focus on ambiguous context, where the questions are clear but depending on which context the models are receiving the answers may vary \cite{neeman-etal-2023-disentqa, Bai_2023, yu-etal-2023-crepe}. However, none of the existing work 1) enable their models to also cite their sources, which is crucial for factuality verification; and 2) evaluate a more complex, real-world setting, which requires multi-hop reasoning.

Prior work on QA in knowledge conflicts settings falls into two categories: ambiguous questions, where answers vary depending on the question phrasing \cite{min2020ambigqa, sun-etal-2023-answering, cole2023selectively}, and ambiguous context, where answers vary depending on the context provided \cite{neeman-etal-2023-disentqa, Bai_2023, yu-etal-2023-crepe}. However, existing work lacks two key aspects: 1) source citation, and 2) evaluation in complex, real-world settings requiring multi-hop reasoning.

\paragraph{Citations}
While LLMs contain various types of factual knowledge within their parameters \cite{shin-etal-2020-autoprompt, shaier-etal-2023-emerging, kassner-schutze-2020-negated, shaier-etal-2024-comparing, sung-etal-2021-language, petroni}, they
often suffer from hallucinations, generating non-factual text \cite{rawte-etal-2023-troubling, semnani-etal-2023-wikichat, li-etal-2023-halueval, dziri-etal-2022-origin}. To address this, researchers have developed models that can provide citations \cite{shaier2023stochasticparrotsimitatingtell, khalifa2024sourceaware, slobodkin2024attribute, bohnet2023attributed, gao-etal-2023-rarr}. However, these models have not been tested in complex scenarios, such as ambiguous QA settings with multiple possible answers or multi-hop reasoning.

% Another issue that LLMs have shown to have is hallucinations \cite{rawte-etal-2023-troubling, semnani-etal-2023-wikichat, li-etal-2023-halueval, dziri-etal-2022-origin}. That is, they produce text that is not factual. While an increasing number of works focus on developing models that can produce citations \cite{shaier-etal-2023-stochastic, khalifa2024sourceaware, slobodkin2024attribute, bohnet2023attributed, gao-etal-2023-rarr}, none are tested on ambiguous QA settings, where multiple answers exist, or multi-hop reasoning.

\section{Experiments}
\label{Experiments}
% In our experiments, each dataset consists of question, contexts, and answers triples $(q,[c_1, c_2, ... , c_n], [a_1, a_2, ... , a_k])$, where $q \in Q$, $c_i \in C$, $a_j \in A$, and $Q$, $C$, and $A$ are collections of questions, contexts, and answers, respectively. More specifically, each question $q$ is accompanied by multiple context documents $[c_1, c_2, ... , c_n]$, and at least two conflicting answers $[a_1, a_2, ... , a_k]$. Following \citet{NEURIPS2020_1457c0d6, chowdhery2022palm, liu2023evaluating}, we concatenate the question and the corresponding contexts into a single string which is sent to each LLM.

In our experiments, each dataset consists of triples $(q, [c_1,..., c_n], [a_1,..., a_k])$, where $q$ is a question, $[c_1,..., c_n]$ are multiple context documents, and $[a_1,..., a_k]$ are at least two conflicting answers. We follow prior work \cite{NEURIPS2020_1457c0d6, chowdhery2022palm, shaier-etal-2024-say, liu2023evaluating} by concatenating the question and contexts into a single string, which is then input to each model.

% \citet{nori2023capabilities, chowdhery2022palm, chung2022scaling, NEURIPS2020_1457c0d6, cheng2023adapting, liu2023evaluating}, 

% :

% \begin{center}
% \texttt{
% Question: <q> Context: <$[c_1, c_2, ... , c_n]$>,   
% }
% \end{center}

% \noindent
% where \texttt{<q>} and each $c_i$ are the question and contexts strings, respectively.
% An example can be seen in Figure \ref{main_figure}.

% As existing QA datasets do not contain citation meta-data, 

\subsection{Metrics}
% We evaluate each model's ability to generate a distinct response for each conflicting piece of information while factually citing their sources. More concretely, for each question $q$, we evaluate the generated response along two dimensions: 1) \textbf{Acc\_K}: the average accuracy of producing at least K of the gold answers. For example, if the gold answers are [“X”, “Y”, “Z”] and the generated answers are [“X”, “Y”], the scores will be: Acc\_1=1, Acc\_2=1, Acc\_3=0; and 2) \textbf{Citation Accuracy}: the average accuracy of producing the citation strings with the correct citation. For example, if the gold answers are [“According to Document X the answer is X1”, “According to Document Y the answer is Y1”] and the generated answers are [“According to Document X the answer is X1”, “According to Document Z the answer is Y1”], the score will be $0.5$. For both accuracy measures, we follow \citet{liu2023lost, mallen-etal-2023-trust, kandpal2023large} and evaluate if the gold answer or citation string are present in the predicted output.

To comprehensively assess the performance of systems tackling the novel task of QA with source citation in ambiguous settings, we introduce two novel evaluation metrics that capture the ability to generate distinct responses for conflicting information while accurately citing sources. Specifically, for each question $q$, we evaluate the generated response along two crucial dimensions:

\textbf{Acc\_K}: This metric measures the ability to produce a diverse set of correct answers, with a focus on generating at least $K$ of the gold answers. For instance, if the gold answers are [“X”, “Y”, “Z”] and the generated answers are [“X”, “Y”], the scores would be: Acc\_1=1, Acc\_2=1, Acc\_3=0.

\textbf{Citation Accuracy (A\_C)}: This metric assesses the ability to accurately generate citation strings corresponding to the correct sources. For example, if the gold answers are [“According to Document X the answer is X1”, “According to Document Y the answer is Y1”] and the generated answers are [“According to Document X the answer is X1”, “According to Document Z the answer is Y1”], the score would be 0.5.

By utilizing these two metrics, we can gain a more nuanced understanding of system performance in resolving knowledge conflicts and citing sources accurately, ultimately driving progress in this critical task. For both accuracy measures, we follow \citet{liu2023lost, mallen-etal-2023-trust, kandpal2023large} and evaluate if the gold answer or citation string are present in the output.

% \subsubsection{Human Evaluation}
% As no automatic evaluation method is perfect, we also manually analyze a subset of 100 questions and their corresponding generated answers from the best performing model and its best performing method (see Section \ref{methods}) for both accuracy measures on each of the datasets (see Section \ref{datasets}).
% Additionally, we More concretely, we evaluate: 1) if the string matching is good; and 2) if the model generate more answers

\subsection{Datasets}
% We assess our methods on three datasets, which are focused on different conflicting settings. We augment the evaluated datasets with citations meta-data. In particular, we add the citation string “Document X” before each document context $c_i$, where X is a unique integer representing each document (see Figure \ref{main_figure} for example). Realistically, the citation string can be anything from a PubMed or a WikiPedia ID corresponding to the document. While one such citation can exist for each full document, for longer contexts (e.g., multi-hop setting) we add the citation string before each paragraph, which increases the task complexity for models as there are more citations to reason through and produce, but also increases the ease of users of finding the exact relevant information as they do not need to parse through the entire document.

% “X”

Notably, existing QA datasets lack citation metadata, which is a critical component of our proposed task. To address this gap, we augment three reading comprehension (RC) datasets to create novel evaluation sets\footnote{Which we will make publicly available.} that focus on different conflicting settings, each enriched with citation metadata. Specifically, we add a unique citation string “Document X” before each document context $c_i$, where X represents a distinct document identifier (as illustrated in Figure \ref{main_figure}). In real-world scenarios, these citation strings can correspond to PubMed IDs, Wikipedia IDs, or other types of document identifiers. To further increase the task's complexity and realism, we add citation strings before each paragraph in longer contexts, such as multi-hop settings. This design choice presents a dual benefit: models must now reason through and produce multiple citations, while users can more easily identify relevant information without having to parse entire documents. Dataset examples can be seen in Table \ref{generation_examples_table}.

\label{datasets}
\paragraph{AmbigQA-Cite}
\label{AmbigQA}
We build upon AmbigQA \cite{min2020ambigqa}, an open-domain RC dataset, which is derived from the Natural Questions (NQ) dataset \cite{kwiatkowski-etal-2019-natural} and comprises 14,042 questions. Notably, AmbigQA-Cite features \textit{ambiguous questions}.
% , each associated with multiple answers. 
To create our citation-augmented dataset, we employ the following methodology: for each ambiguous question, we select contexts that contain exactly one of the answers and exclude those that contain multiple answers. We further restrict our dataset to questions with exactly two conflicting answers, each supported by a distinct context, as questions with more conflicting answers are extremely rare and would lead to limited sample sizes and unreliable conclusions. The resulting dataset, which we term \textit{AmbigQA-Cite}, is enriched with citation information to support the development of more accurate and trustworthy question answering models.

% AmbigQA-Cite \cite{min2020ambigqa} is an open-domain, RC QA dataset that contains 14,042 questions from the Natural Questions (NQ) dataset \cite{kwiatkowski-etal-2019-natural}. AmbigQA-Cite contains \textit{ambiguous questions}, each corresponding to multiple answers. We utilize it as follows: for each ambiguous question, we select contexts that each contain exactly one of the answers and filter those that contain more. We only use questions that contain two conflicting answers in two distinct contexts, as questions with more conflicting answer are extremely rare, making it challenging to derive meaningful conclusions due to the limited sample size. We term the citation-augmented dataset AmbigQA-Cite.

% ; and 2) questions with a higher number of answers introduce a much larger number of contexts, surpassing the maximum context size manageable by the models.
% , which can be seen in our results where models' performance often . 
% This limitation is evident in our methodology, where we can only utilize a limited range of examples (1-5) in a few-shot scenario. Additionally, Table xxx illustrates that the model's performance deteriorates significantly when the number of shots exceeds three.
% as the number of questions that contain more answers is extremely low and will prevent us from drawing conclusions for the small sample size. Additionally,  

\paragraph{DisentQA-DupliCite}
\label{DisentQA}

We use DisentQA \cite{neeman-etal-2023-disentqa}, an open-domain RC QA dataset with 108,291 questions from the NQ dataset. Unlike AmbigQA, DisentQA focuses on \textit{ambiguous contexts}, where the question is clear, but the answer varies depending on the context (Figure \ref{main_figure}). The dataset uses entity-substitution \cite{longpre-etal-2021-entity} to create conflicting contexts, resulting in 39,716 pairs of questions with two conflicting contexts and answers each. Notably, this substitution approach leads to context duplication, where both contexts for each question are similar except for the replaced entity. We augment this dataset with citation information, creating \textit{DisentQA-DupliCite}.

\paragraph{DisentQA-ParaCite}
To mitigate the potential shortcut issue in \textit{DisentQA-DupliCite}, where models may exploit the similarity between duplicated contexts, we create a paraphrased version of each conflicting context for each question.
% This approach encourages models to focus on the underlying semantic differences between contexts rather than superficial similarities. 
Specifically, we use ChatGPT \cite{openai2023} to paraphrase each conflicting context, taking care to preserve the replaced entity in the output using the specific prompt: “\textit{Paraphrase this: \{conflicting\_context\}. Ensure that \{conflicting\_label\} is still in the paraphrased output}”. This process yields a new dataset, which we term \textit{DisentQA-ParaCite}, featuring paraphrased contexts that require models to engage in more robust and meaningful reasoning.\footnote{We manually evaluate 100 paraphrased examples and found that $98\%$ were of high quality.}

\paragraph{Conflicting HotPotQA-Cite}
\label{hotpot}
% HotPotQA \cite{yang-etal-2018-hotpotqa} is multi-hop RC QA dataset that contains 112,779 questions, where the questions require models to reason over multiple documents to reach an answer. As the original dataset does not contain conflicting contexts, we follow \citet{shaier-etal-2024-desiderata, contra, li-etal-2020-bert-attack}'s approach of using a masked language model (MLM) to augment each context with conflicting answers. We opt for the MLM approach instead of the entity-substitution approach DisentQA use, as the latter can influence the grammatically of the text \cite{eisenstein2022honest}. More concretely, we mask the answer string from the context and use DistilBERT \cite{sanh2020distilbert} to generate two additional conflicting answers that are not the original answer. We then replace the original answer with each of the two conflicting answers to produce three conflicting answers and contexts for each question (i.e., the original plus the two conflicting ones). We call the resulting dataset \textit{Conflicting HotPotQA-Cite}. To the best of our knowledge, this in turn creates the first conflicting multi-hop QA dataset featuring real-world, naturally occurring contexts\footnote{While one multi-hop conflicting QA dataset exist \cite{kazemi2023boardgameqa}, it is composed of isolated facts, which are based on rules, which are not grounded in complex contexts \cite{sprague2023musr}.}.

HotPotQA \cite{yang-etal-2018-hotpotqa} is a multi-hop RC QA dataset with 112,779 questions
% , requiring models to reason across multiple documents. 
We use a masked language model (MLM) approach, similar to \citet{shaier-etal-2024-desiderata, contra, li-etal-2020-bert-attack}, to introduce conflicting contexts. We opt for MLM over entity substitution to preserve text grammatical integrity \cite{eisenstein2022honest}. Using DistilBERT \cite{sanh2020distilbert}, we generate two conflicting answers per context, creating three conflicting answers and contexts per question. This yields the \textit{Conflicting HotPotQA-Cite} dataset, the first conflicting multi-hop QA dataset with real-world, naturally occurring contexts. Unlike BoardgameQA \cite{kazemi2023boardgameqa}, our dataset features complex contextualized contradictions.

% we adopt a masked language model (MLM) approach, similar to \citet{shaier-etal-2024-desiderata, contra, li-etal-2020-bert-attack}. We opt for MLM instead of entity substitution, as the latter can compromise the grammatical integrity of the text \cite{eisenstein2022honest}. Using DistilBERT \cite{sanh2020distilbert}, we generate two conflicting answers for each context, replacing the original answer with each of them to create three conflicting answers and contexts per question. This results in the \textit{Conflicting HotPotQA-Cite} dataset, which, to our knowledge, is the first conflicting multi-hop QA dataset featuring real-world, naturally occurring contexts. Unlike BoardgameQA \cite{kazemi2023boardgameqa}, which contains isolated facts based on rules, our dataset provides complex contextualized contradictions. 

We provide two variants of this dataset: (1) a \textit{with distractors} version, which includes up to 14 cited documents in each context, including both relevant and distracting contexts, and (2) a \textit{no distractors} version, which only includes the relevant contexts, limited to up to 6 cited documents.

\subsection{Models}
\begin{table}[t]
\centering
\small
\setlength{\tabcolsep}{2.0pt}

\begin{tabular}{lll}
\multicolumn{1}{c}{\textbf{Model}} & \multicolumn{1}{c}{\textbf{Paramaters}} & \multicolumn{1}{c}{\textbf{Training Dataset Size}} \\
Llama-7B Chat                     & 7 billion                               & 2 trillion tokens                                  \\
Llama-13B Chat                        & 13 billion                              & 2 trillion tokens                                  \\
Llama-70B Chat                         & 70 billion                              & 2 trillion tokens                                  \\
MPT-7B Instruct                             & 7 billion                               & 1 trillion tokens                                  \\
Falcon-7B                          & 7 billion                               & 1.5 trillion tokens                               
\end{tabular}

\caption{Models, their size, and the number of tokens in their training data.}
\label{models_table}
\end{table}

We experiment with 5 different LLMs: Llama-2-7B Chat \cite{touvron2023llama}, Llama-2-13B Chat \cite{touvron2023llama}, Llama-2-70B Chat\footnote{We evaluate the 70B model on most settings, except a few due to unexpected computational constraints.} Instruct \cite{touvron2023llama}, MPT-7B \cite{mpt_2023}, and Falcon-7B Instruct \cite{almazrouei2023falcon}. A summary can be seen in Table \ref{models_table}. 

\subsection{Baselines}
\label{methods}
% We focus on prompting-type methods, as many previous work have shown their success in improving models' performance without the need for further finetuning \cite{NEURIPS2020_1457c0d6, liu2021pretrain, wei2023chainofthought, dong2023survey}. 

% In addition to proposing the new task of QA with source citation in ambiguous settings, we propose several strong baselines composing of rule, prompting, and finetuning-based models.

In addition to introducing the novel task of QA with source citation in ambiguous settings, we establish a set of strong baseline models to facilitate progress in this area. Our proposed baselines comprise a range of approaches, including rule-based, prompting-based, and finetuning-based models.

In the following examples $q_{e1}$,...,$q_{ek}$ and $c_{e1}$,...,$c_{en}$ are in-context learning question and contexts; $q_{ti}$ is the test question and $c_{ti}$ are the test contexts. The citations are included in the contexts. 

% To clarify our notation, we define the following variables:
% $q_{e1}$,..., $q_{ek}$: in-context learning questions
% $c_{e1}$,..., $c_{en}$: in-context learning contexts, which include citations
% $q_{ti}$: test question
% $c_{ti}$: test contexts, which also include citations
% These notations will be used throughout our experiments and analyses.

\paragraph{Zero-shot Baseline}
We concatenate the question and contexts into a single input string, as described in Section \ref{Experiments}, and feed it to each of the models.

% :

% \begin{center}
% \texttt{
% Question: $q_{ti}$ Context: <$c_{ti}$>. Answer:
% }
% \end{center}
% \begin{center}
% \texttt{
% Question: $q_{ti}$ Context: <$[c_1, c_2, ... , c_n]$>,   
% }
% \end{center}

% \paragraph{Conflict-aware Prompting}

\subsubsection{Prompt-based Methods}
See Appendix \ref{prompt_design} for detailed prompt designs.

\paragraph{Conflict-aware Basic Prompting}
We employ a few-shot approach \cite{NEURIPS2020_1457c0d6} with 1, 3, or 5 examples per prompt, utilizing a structured prompt design that explicitly acknowledges the presence of conflicting information.
This conflict-aware (C.A) prompting design emphasizes the existence of conflicting information and its corresponding citations, enabling models to develop a more nuanced understanding of ambiguous contexts.

\paragraph{Few-shot Conflict-aware CoT Prompting}
We adopt the few-shot Chain-of-Thought (CoT) method \cite{wei2023chainofthought}, which involves providing the model with explicit reasoning steps to arrive at an answer. We create 1 or 3 manually-crafted CoT examples that highlight conflicting information and their associated citations, and append them to the prompt, enabling the model to generate an answer in a single step.

\paragraph{Zero-shot CoT Prompting}
\label{Zero-shot-cot}
In the zero-shot approach \cite{kojima2023large}, we employ a two-step process to elicit reasoning from the model: 
% Specifically:
% we append the string “Let’s think step by step” to the prompt, get a response from the model in step 1, and append that response to the final question and context string in step 2:

Unlike the C.A CoT method, here, we do not provide explicit examples of conflicting context and citations. 
% in this zero-shot approach. 
Instead, we aim to assess whether the model's self-generated reasoning paths are sufficient to handle conflicting facts.

\subsubsection{Rule-based Methods}
\paragraph{Document Split}
\label{doc_split}
% Here, for each question $q$, instead of concatenating all documents together into a single string we send each question-context pair to the models separately: “Question: <q> Context: <c1>” , “Question: <q> Context: <c2>”, [...]. This approach has the advantage of making citations trivial, as the models generate one response for each document, which we evaluate separately. However, as models can only see one document at a time they are incapable of answering questions that require complex reasoning across documents. 

Our rule-based approach, \textit{Document Split}, employs a predetermined set of rules to process the context. Specifically, we split the context into individual articles based on the citation tokens, and process them sequentially, following a strict rule: each article is processed one at a time, rather than all at once. This approach makes citations trivial, as we can generate one response per document and evaluate them separately to identify correct citations.
% This approach has a significant advantage, as it renders citations trivial. By generating one response for each document, we can evaluate them separately, making it easier to identify the correct citations. 
However, this rule-based approach also has a limitation. Since models can only see one document at a time, they are incapable of answering questions that require complex reasoning across multiple documents.

% Rule-Based Aspects Splitting the Articles: Deciding to split the context into individual articles and processing them sequentially follows a predetermined rule. Sequential Processing: The rule specifies that each article should be processed one at a time rather than all at once. Concatenation of Answers: After obtaining the answers, combining them into a final response is also governed by a specific rule.

% This approach simulate a perfect retrieval system which disambiguate the question or context and only provides the relevant information. 

% However, while it is feasible for users to send one document at a time to models, in practice the contexts are normally concatenated into one string for ease, and also since many questions require reasoning over multiple documents which would be impossible in this approach.

\subsubsection{Finetuning Methods}
\paragraph{Fine-tuning with Low-Rank Adaptation (LoRA)}
We fine-tune LLMs on our datasets using LoRA \cite{hu2021lora}, a parameter-efficient technique that avoids full model fine-tuning. LoRA adds small, trainable adapters to specific layers, keeping original parameters frozen, and allows control over adapter influence via the alpha value.
% We utilize LoRA \cite{hu2021lora}, a parameter-efficient technique, to fine-tune LLMs on our datasets. This approach avoids the computationally expensive and time-consuming process of fine-tuning the entire base model. Instead, LoRA adds a small set of trainable parameters to the model, while keeping the original model parameters frozen. Specifically, we insert adapters into specific layers, and only update the added weights during training, allowing us to control the influence of these adapters through the alpha value, which determines the weight assigned to the adapters. 
We fine-tune each model on each of the following datasets: AmbigQA-Cite, DisentQA-DupliCite, and Conflicting HotPotQA-Cite (without distractors). The complete set of hyperparameters used for fine-tuning can be found in Appendix \ref{finetuning_parap}.

% \paragraph{Llama-2-7B} Llama-2-7B \cite{touvron2023llama} is a LLM which contains 7 billion parameters and is trained on 2 trillion tokens. 

% \paragraph{Llama-2-13B} Llama-2-13B \cite{touvron2023llama} is a LLM which contains 13 billion parameters and is trained on 2 trillion tokens. 

% \paragraph{Llama-2-70B} Llama-2-70B \cite{touvron2023llama} is a LLM which contains 70 billion parameters and is trained on 2 trillion tokens. 

% \paragraph{MPT-7B} MPT-7B \cite{mpt_2023} is a LLM which contains 7 billion parameters and is trained on 1 trillion tokens. 

% \paragraph{Falcon-7B} Falcon-7B \cite{almazrouei2023falcon} is a LLM which contains 7 billion parameters and is trained on 1.5 trillion tokens. 

\section{Results}
\begin{table*}[t]
\centering
\tiny
\setlength{\tabcolsep}{0.3pt}

\renewcommand{\arraystretch}{1}
\begin{tabular}{ccccc}
\textbf{Dataset}          & \textbf{Question}                                                                                                                                 & \textbf{Context}                                                                     & \textbf{\begin{tabular}[c]{@{}c@{}}Zero-shot\\ Answer\end{tabular}}                                                                                                                   & \textbf{\begin{tabular}[c]{@{}c@{}}Few-shot\\ Answer\end{tabular}}                                                                                                                                                                                                                                                 \\ \hline
\multirow{2}{*}{AmbigQA-Cite}  & \begin{tabular}[c]{@{}c@{}}Who played warden\\ hodges in dad's army?\end{tabular}                                                                 & \begin{tabular}[c]{@{}c@{}}{}[...] \textbf{Bill Pertwee} as Hodges [...] \textbf{Martin} \\ \textbf{Savage} is an English film, stage and \\ television actor. He appeared in both series \\[0.1cm]\end{tabular}      & \begin{tabular}[c]{@{}c@{}}Bill Pertwee played\\ Warden Hodges\\ in Dad's Army\end{tabular}                                                                                           & \begin{tabular}[c]{@{}c@{}}According to Document 1 the answer is: \\Bill Pertwee.\\ According to Document 2 the answer is: \\Martin Savage\end{tabular}                                                                                                                                                                \\
                          & \begin{tabular}[c]{@{}c@{}}Who voices rocket raccoon\\ in guardians of the galaxy?\end{tabular}                                                   & \begin{tabular}[c]{@{}c@{}}Disney Studios [...] and \textbf{Bradley Cooper} \\ as the titular Guardians [...] This season \\ included [...] \textbf{Trevor Devall} - Rocket Raccoon \end{tabular}    & \begin{tabular}[c]{@{}c@{}}Trevor Devall voices\\ Rocket Raccoon in\\ Guardians of the Galaxy\end{tabular}                                                                            & \begin{tabular}[c]{@{}c@{}}According to Document 1 the answer is: \\Bradley Cooper.\\ According to Document 2 the answer is: \\Trevor Devall\end{tabular}                                                                                                                                                              \\ \hline
\multirow{2}{*}{DisentQA-DupliCite} & \begin{tabular}[c]{@{}c@{}}who gave the idea of\\ separate independent\\ muslim state in india\end{tabular}                                       & \begin{tabular}[c]{@{}c@{}}[...] The Aligarh Muslim University in \\ which \textbf{Syed Ahmad Khan} was a central \\ figure [...] The Aligarh Muslim University in \\ which \textbf{Joe Kennedy III} was a central figure \\[0.1cm]
\end{tabular} & \begin{tabular}[c]{@{}c@{}}The idea of a separate \\ independent Muslim state \\ in India was given \\ by Syed Ahmad Khan {[}...{]}\end{tabular}                                         & \begin{tabular}[c]{@{}c@{}}According to Document 1 the answer is: \\Syed Ahmad Khan.\\ According to Document 2 the answer is: \\Joe Kennedy III\end{tabular}                                                                                                                                                           \\
                          & \begin{tabular}[c]{@{}c@{}}who gave the first in\\ person state of the union\end{tabular}                                                         & \begin{tabular}[c]{@{}c@{}} [...] After 1913, \textbf{Woodrow Wilson}, \\ the 28th U.S. President [...] \\ After 1913, \textbf{Angela Hunte}, \\ the 28th U.S. President [...] \\[0.1cm]
                          \end{tabular}     & \begin{tabular}[c]{@{}c@{}}Woodrow Wilson was the first\\ president to deliver {[}...{]}\end{tabular}                                                                                 & \begin{tabular}[c]{@{}c@{}}According to Document 1 the answer is: \\Woodrow Wilson.\\ According to Document 2 the answer is: \\Angela Hunte\end{tabular}                                                                                                                                                               \\
                          
                          \hline
\multirow{2}{*}{Conflicting HotpotQA-Cite}  
                          & \begin{tabular}[c]{@{}c@{}}China Bio-Immnunity has\\ developed a vaccine for\\ which viral disease that\\ causes brain inflammation?\end{tabular} & \begin{tabular}[c]{@{}c@{}}
                          [...] \textbf{Ebola} is a viral disease that causes \\ inflammation of the brain [...]
                          \\ \textbf{Rabies} is a viral disease that causes \\ inflammation of the brain [...]
                          \\ \textbf{Zika} is a viral disease that causes \\ inflammation of the brain [...]
                          \end{tabular}          & \begin{tabular}[c]{@{}c@{}}China Bio-Immunity has \\ developed a vaccine for \\ Ebola [...]\end{tabular}                         & \begin{tabular}[c]{@{}c@{}}According to Documents \\{[}'Document 3', 'Document 7'{]}\\ the answer is Rabies.\\ According to Documents \\{[}'Document 10', 'Document 12'{]}\\ the answer is Ebola.\\ According to Documents \\{[}'Document 11', 'Document 14'{]}\\ the answer is Zika.\end{tabular}                       \\
                          & \begin{tabular}[c]{@{}c@{}}Who did the player nicknamed\\ "The Human Highlight\\ Film" play for after he\\ left the Atlanta Hawks?\end{tabular}   & \begin{tabular}[c]{@{}c@{}}
                          [...] who signed with the \textbf{Heat} in the \\ offseason [...] who signed with the \\ \textbf{Hawks} in the offseason
                          [...] who \\ signed with the \textbf{Boston Celtics} in the \\ offseason
                          
                          \end{tabular}  & \begin{tabular}[c]{@{}c@{}}The player nicknamed\\ "The Human Highlight\\ Film" (Dominique Wilkins)\\ played for the Boston \\ Celtics after he left the\\ Atlanta Hawks.\end{tabular} & \begin{tabular}[c]{@{}c@{}}According to Documents \\{[}'Document 6', 'Document 7'{]}\\ the answer is Boston Celtics.\\ According to Documents \\{[}'Document 11', 'Document 13'{]}\\ the answer is Atlanta Hawks.\\ According to Documents \\{[}'Document 10', 'Document 14'{]}\\ the answer is Miami Heat.\end{tabular} \\ 
                          \hline

\multirow{3}{*}{DisentQA-ParaCite} & \begin{tabular}[c]{@{}c@{}}who gave the idea of\\ separate independent\\ muslim state in india\end{tabular}                                       & \begin{tabular}[c]{@{}c@{}}[...] The Aligarh Muslim University in \\ which \textbf{Syed Ahmad Khan} was a central \\ figure [...] The Aligarh Muslim University \\ where \textbf{Joe Kennedy} III played a key role \\[0.1cm]
\end{tabular} & \begin{tabular}[c]{@{}c@{}}The concept of an \\ independent Muslim state \\ within India \\ was proposed by Syed \\ Ahmad Khan. {[}...{]}\end{tabular}                                         & \begin{tabular}[c]{@{}c@{}}According to Document 1 the answer is: \\Syed Ahmad Khan.\\ According to Document 2 the answer is: \\Joe Kennedy III\end{tabular}                                                                                                                                                           \\
                          & \begin{tabular}[c]{@{}c@{}}who gave the first in\\ person state of the union\end{tabular}                                                         & \begin{tabular}[c]{@{}c@{}} [...] After 1913, \textbf{Woodrow Wilson}, \\ the 28th U.S. President [...] \\ Following 1913, \textbf{Angela Hunte}, \\ the 28th President of \\ the United States \\[0.1cm]
                          \end{tabular}     & \begin{tabular}[c]{@{}c@{}}Woodrow Wilson was the first\\ president to deliver {[}...{]}\end{tabular}                                                                                 & \begin{tabular}[c]{@{}c@{}}According to Document 1 the answer is: \\Woodrow Wilson.\\ According to Document 2 the answer is: \\Angela Hunte\end{tabular}                                                                                                                                                                                                                                                                      \\ \hline

\end{tabular}

\caption{Generation Examples for Llama-70B (3-Shot Setting).}
\label{generation_examples_table}
\end{table*}
% For illustrative purposes, example model generations for all datasets are provided in Table \ref{generation_examples_table}.
Example model generations are shown in Table \ref{generation_examples_table}.

\subsection{Ambiguous Questions}
\label{Ambiguous_questions}
\begin{table*}[t]
\centering
\small
\setlength{\tabcolsep}{2.0pt}
\begin{tabular}{|c|l|lll|ll|l|l|l|}
\hline
\begin{tabular}[c]{@{}c@{}}\textbf{Method} /\\ \textbf{Model}\end{tabular} & \multicolumn{1}{c|}{\textbf{Zero-Shot}}                                                                                                           & \multicolumn{3}{c|}{
% \begin{tabular}[c]{@{}c@{}}\textbf{Conflict-aware} \\ \textbf{Basic Prompting}\end{tabular}
\textbf{C.A Basic Prompting}
}                                                                                                                                                                                                                                                                                                                                                                                                                                                                    & \multicolumn{2}{c|}{\begin{tabular}[c]{@{}c@{}}\textbf{Few-shot} \\ \textbf{C.A CoT}\end{tabular}}                                                                                                                                                                                                                                                                                           & \multicolumn{1}{c|}{\begin{tabular}[c]{@{}c@{}}\textbf{Zero-shot} \\ \textbf{CoT}\end{tabular}}                                                                                                            & \multicolumn{1}{c|}{\begin{tabular}[c]{@{}c@{}}\textbf{Document} \\ \textbf{Split}\end{tabular}}                                                                                                       & \multicolumn{1}{c|}{\textbf{Finetuning}} \\ \hline
                                                        & \multicolumn{1}{c|}{}                                                                                                                             & \multicolumn{1}{c|}{\textbf{1-shot}}                                                                                                                                     & \multicolumn{1}{c|}{\textbf{3-shot}}                                                                                                                                     & \multicolumn{1}{c|}{\textbf{5-shot}}                                                                                                                & \multicolumn{1}{c|}{\textbf{1-shot}}                                                                                                                                       & \multicolumn{1}{c|}{\textbf{3-shot}}                                                                                                                  & \multicolumn{1}{c|}{\textbf{1-shot}}                                                                                                                & \multicolumn{1}{c|}{\textbf{}}                                                                                                                      & \multicolumn{1}{c|}{}                                                                                                                              \\ \hline
Llama-7B                                                & \begin{tabular}[c]{@{}l@{}}A\_1: 54.8  \\ A\_2: 2.1  \\ A\_C: 0.0\end{tabular} & \multicolumn{1}{l|}{\begin{tabular}[c]{@{}l@{}}A\_1: 54.8  \\ A\_2: 20.4  \\ A\_C: 33.3\end{tabular}} & \multicolumn{1}{l|}{\begin{tabular}[c]{@{}l@{}}A\_1: 62.3  \\ A\_2: 21.5  \\ A\_C: 34.4\end{tabular}} & \begin{tabular}[c]{@{}l@{}}A\_1: 61.2  \\ A\_2: 24.7  \\ A\_C: 34.9\end{tabular} & \multicolumn{1}{l|}{\begin{tabular}[c]{@{}l@{}}A\_1: 41.9  \\ A\_2: 8.6  \\ A\_C: 2.1\end{tabular}}  & \begin{tabular}[c]{@{}l@{}}A\_1: 56.9 \\ A\_2: 13.9  \\ A\_C: 0.5\end{tabular}  & \begin{tabular}[c]{@{}l@{}}A\_1: 45.1 \\ A\_2: 2.1  \\ A\_C: 0.0\end{tabular} & \begin{tabular}[c]{@{}l@{}} A\_1: 67.7 \\ A\_2: 30.1  \\ A\_C: NA\end{tabular} & \begin{tabular}[c]{@{}l@{}}A\_1: \textbf{69.8}  \\ A\_2: \textbf{35.4}  \\  A\_C: \textbf{48.3} \end{tabular} \\ \hline
Llama-13B                                               & \begin{tabular}[c]{@{}l@{}}A\_1: 48.3  \\ A\_2: 3.2  \\ A\_C: 0.0\end{tabular} & \multicolumn{1}{l|}{\begin{tabular}[c]{@{}l@{}}A\_1: 63.4  \\ A\_2: 23.6  \\ \textbf{A\_C: 36.5}\end{tabular}} & \multicolumn{1}{l|}{\begin{tabular}[c]{@{}l@{}}\textbf{A\_1: 67.7}  \\ A\_2: 22.5  \\ \textbf{A\_C: 36.5}\end{tabular}} & \begin{tabular}[c]{@{}l@{}}A\_1: 63.4  \\ A\_2: 23.6  \\ A\_C: 34.9\end{tabular} & \multicolumn{1}{l|}{\begin{tabular}[c]{@{}l@{}}A\_1: 61.2 \\ A\_2: 10.7  \\ A\_C: 5.9\end{tabular}}  & \begin{tabular}[c]{@{}l@{}}A\_1: 55.9 \\ A\_2: 15.0  \\ A\_C: 9.6\end{tabular}  & \begin{tabular}[c]{@{}l@{}}A\_1: 45.1 \\ A\_2: 1.0  \\ A\_C: 0.0\end{tabular} & \begin{tabular}[c]{@{}l@{}}A\_1: 62.3 \\ A\_2: 21.5  \\ A\_C: NA\end{tabular} & \begin{tabular}[c]{@{}l@{}}A\_1: 66.6  \\ A\_2: \textbf{32.5}  \\  A\_C: 30.6 \end{tabular} \\ \hline
Llama-70B                                               & \begin{tabular}[c]{@{}l@{}}A\_1: 54.8  \\ A\_2: 4.3  \\ A\_C: 0.0\end{tabular} & \multicolumn{1}{l|}{\begin{tabular}[c]{@{}l@{}}A\_1: 72.0  \\ \textbf{A\_2: 35.4}  \\ A\_C: 45.6\end{tabular}} & \multicolumn{1}{l|}{\begin{tabular}[c]{@{}l@{}}A\_1: 74.1  \\ \textbf{A\_2: 35.4}  \\ \textbf{A\_C: 48.3}\end{tabular}} & \begin{tabular}[c]{@{}l@{}}A\_1: 70.9  \\ A\_2: 30.1  \\ A\_C: 45.6\end{tabular}                                                                                                                                                 & \multicolumn{1}{l|}{\begin{tabular}[c]{@{}l@{}}A\_1: 70.9 \\ A\_2: 30.1  \\ A\_C: 29.0\end{tabular}} & \begin{tabular}[c]{@{}l@{}}A\_1: 73.1 \\ A\_2: 31.1  \\ A\_C: 31.7\end{tabular} & \begin{tabular}[c]{@{}l@{}}A\_1: 38.7 \\ A\_2: 4.3  \\ A\_C: 0.0\end{tabular} & \begin{tabular}[c]{@{}l@{}}\textbf{A\_1: 76.3} \\ A\_2: 25.8  \\ A\_C: NA\end{tabular} & \begin{tabular}[c]{@{}l@{}}A\_1: -  \\ A\_2: -  \\  A\_C: -\end{tabular} \\ \hline
MPT-7B                                                  & \begin{tabular}[c]{@{}l@{}}A\_1: 50.5  \\ A\_2: 0.0  \\ A\_C: 0.0\end{tabular} & \multicolumn{1}{l|}{\begin{tabular}[c]{@{}l@{}}A\_1: 51.6  \\ A\_2: 9.6  \\ A\_C: 12.9\end{tabular}}  & \multicolumn{1}{l|}{\begin{tabular}[c]{@{}l@{}}A\_1: 46.2  \\ A\_2: 9.6  \\ \textbf{A\_C: 21.5}\end{tabular}}  & \begin{tabular}[c]{@{}l@{}}A\_1: 44.0  \\ A\_2: 7.5  \\ A\_C: 19.8\end{tabular}  & \multicolumn{1}{l|}{\begin{tabular}[c]{@{}l@{}}A\_1: 47.3 \\ A\_2: 3.2  \\ A\_C: 8.6\end{tabular}}   & \begin{tabular}[c]{@{}l@{}}A\_1: 45.1 \\ A\_2: 2.1  \\ A\_C: 6.9\end{tabular}   & \begin{tabular}[c]{@{}l@{}}A\_1: 45.1 \\ A\_2: 1.0  \\ A\_C: 0.0\end{tabular} & \begin{tabular}[c]{@{}l@{}}\textbf{A\_1: 65.5} \\ \textbf{A\_2: 21.5}  \\ A\_C: NA\end{tabular} & \begin{tabular}[c]{@{}l@{}}A\_1: 51.6  \\ A\_2: 10.7  \\  A\_C: 16.1\end{tabular} \\ \hline
Falcon-7B                                               & \begin{tabular}[c]{@{}l@{}}A\_1: 30.1  \\ A\_2: 1.0  \\ A\_C: 0.0\end{tabular} & \multicolumn{1}{l|}{\begin{tabular}[c]{@{}l@{}}A\_1: 8.6  \\ A\_2: 2.1  \\ A\_C: 4.8\end{tabular}}    & \multicolumn{1}{l|}{\begin{tabular}[c]{@{}l@{}}A\_1: 39.7  \\ A\_2: 5.3  \\ \textbf{A\_C: 16.6}\end{tabular}}  & \begin{tabular}[c]{@{}l@{}}A\_1: 25.8  \\ A\_2: 4.2  \\ A\_C: 8.0\end{tabular}   & \multicolumn{1}{l|}{\begin{tabular}[c]{@{}l@{}}A\_1: 26.8 \\ A\_2: 3.2  \\ A\_C: 11.2\end{tabular}}  & \begin{tabular}[c]{@{}l@{}}A\_1: 36.5 \\ A\_2: 3.2  \\ A\_C: 13.9\end{tabular}  & \begin{tabular}[c]{@{}l@{}}A\_1: 30.1 \\ A\_2: 1.0  \\ A\_C: 0.0\end{tabular} & \begin{tabular}[c]{@{}l@{}}\textbf{A\_1: 52.6} \\ A\_2: 9.6  \\ A\_C: NA\end{tabular}  & \begin{tabular}[c]{@{}l@{}}A\_1: 48.3  \\ A\_2: \textbf{19.3}  \\  A\_C: 13.9 \end{tabular} \\ \hline
\end{tabular}
\caption{AmbigQA-Cite Results. Accuracy scores are reported as percentages. The Document Split method involves providing each document individually to the models, and hence, citations are known by default. C.A=Conflict-aware.}

% A\_1=Average accuracy when at least one of the two answers exist in the generated text; A\_2=Average accuracy when both of the answers exist in the generated text; A\_C=Average citation accuracy. 
\label{ambigqa_table}
\end{table*}
We first analyze the ability of models to answer ambiguous questions on the AmbigQA-Cite dataset
% (see Section \ref{AmbigQA}). 
Results can be seen in Table \ref{ambigqa_table}.

Our analysis of the zero-shot baselines reveals that most models can answer at least one of the two answers correctly (A\_1) around $50\%$ of the time, with Llama-70B performing the best at $54.8\%$ and Falcon-7B performing the worst at $30.1\%$. However, all models struggle to produce distinct answers, with the best A\_2 score being $4.3\%$ for Llama-70B. Moreover, none of the
% out-of-the-box 
models generate citations, resulting in $0\%$ A\_C across all models.

The various prompting methods show improvement in models' ability to answer at least one answer correctly (A\_1), with the best method -- C.A basic -- yielding the highest increase in performance, on Llama-13B with a $19.4\%$ increase. Almost all methods, except for the zero-shot CoT, also improve models' ability to generate distinct responses, with the finetuning method showing the highest increase in A\_2 accuracy, on Llama-7B with a $33.3\%$ increase. In contrast, the zero-shot CoT method performs poorly, with most models and metrics showing a decrease in performance.

The document split method improves all models' A\_1 scores, but not always their A\_2 scores, where
% , and has limitations for questions requiring reasoning over multiple documents. 
finetuning results are mixed, with some models (like Llama-7B) outperforming the best prompting method, while others (like Llama-70B) show comparable or weaker performance.

\subsection{Ambiguous Context: Single-hop}
\label{Ambiguous_Context_Single}
\begin{table*}[t]
\centering
\small
\setlength{\tabcolsep}{2.0pt}
\begin{tabular}{|c|l|lll|ll|l|l|l|}
\hline
\begin{tabular}[c]{@{}c@{}}\textbf{Method} /\\ \textbf{Model}\end{tabular} & \multicolumn{1}{c|}{\textbf{Zero-Shot}}                                                                                                           & \multicolumn{3}{c|}{
% \begin{tabular}[c]{@{}c@{}}\textbf{Conflict-aware} \\ \textbf{Basic Prompting}\end{tabular}
\textbf{C.A Basic Prompting}
}                                                                                                                                                                                                                                                                                                                                                                                                                                                                         & \multicolumn{2}{c|}{\begin{tabular}[c]{@{}c@{}}\textbf{Few-shot} \\ \textbf{C.A CoT}\end{tabular}}                                                                                                                                                                                                                                                                                           & \multicolumn{1}{c|}{\begin{tabular}[c]{@{}c@{}}\textbf{Zero-shot} \\ \textbf{CoT}\end{tabular}}                                                                                                            & \multicolumn{1}{c|}{\begin{tabular}[c]{@{}c@{}}\textbf{Document} \\ \textbf{Split}\end{tabular}}                                                                                                       & \multicolumn{1}{c|}{\textbf{Finetuning}}                                                                                                       \\ \hline
                                                        & \multicolumn{1}{c|}{}                                                                                                                             & \multicolumn{1}{c|}{\textbf{1-shot}}                                                                                                                                     & \multicolumn{1}{c|}{\textbf{3-shot}}                                                                                                                                     & \multicolumn{1}{c|}{\textbf{5-shot}}                                                                                                                & \multicolumn{1}{c|}{\textbf{1-shot}}                                                                                                                                       & \multicolumn{1}{c|}{\textbf{3-shot}}                                                                                                                  & \multicolumn{1}{c|}{\textbf{1-shot}}                                                                                                                & \multicolumn{1}{c|}{\textbf{}}                                                                                                                      & \multicolumn{1}{c|}{}                                                                                                                      \\ \hline

Llama-7B                                                & \begin{tabular}[c]{@{}l@{}}A\_1: 84.6  \\ A\_2: 10.2  \\ A\_C: 0.0\end{tabular} & \multicolumn{1}{l|}{\begin{tabular}[c]{@{}l@{}}A\_1: 85.9  \\ A\_2: 64.0  \\ A\_C: 51.6\end{tabular}} & \multicolumn{1}{l|}{\begin{tabular}[c]{@{}l@{}}\textbf{A\_1: 88.5}  \\ \textbf{A\_2: 76.4}  \\ \textbf{A\_C: 77.6}\end{tabular}} & \begin{tabular}[c]{@{}l@{}}A\_1: 0.1  \\ A\_2: 0.0  \\ A\_C: 0.0\end{tabular} & \multicolumn{1}{l|}{\begin{tabular}[c]{@{}l@{}}A\_1: 81.7  \\ A\_2: 51.4  \\ A\_C: 14.4\end{tabular}} & \begin{tabular}[c]{@{}l@{}}A\_1: 86.3  \\ A\_2: 68.7  \\ A\_C: 50.0\end{tabular} & \begin{tabular}[c]{@{}l@{}}A\_1: 81.5  \\ A\_2: 14.9  \\ A\_C: 0.0\end{tabular} & \begin{tabular}[c]{@{}l@{}}A\_1: 87.5  \\ A\_2: 49.0  \\ A\_C: NA\end{tabular} & \begin{tabular}[c]{@{}l@{}}A\_1: 79.3  \\ A\_2: 61.0  \\  A\_C: 58.5 \end{tabular}  \\ \hline
Llama-13B                                               & \begin{tabular}[c]{@{}l@{}}A\_1: 82.2  \\ A\_2: 10.5  \\ A\_C: 0.0\end{tabular} & \multicolumn{1}{l|}{\begin{tabular}[c]{@{}l@{}}A\_1: 89.0  \\ A\_2: 74.5  \\ A\_C: 76.0\end{tabular}} & \multicolumn{1}{l|}{\begin{tabular}[c]{@{}l@{}}\textbf{A\_1: 91.9}  \\ \textbf{A\_2: 79.0}  \\ \textbf{A\_C: 81.9}\end{tabular}} & \begin{tabular}[c]{@{}l@{}}A\_1: 0.1  \\ A\_2: 0.0  \\ A\_C: 0.0\end{tabular} & \multicolumn{1}{l|}{\begin{tabular}[c]{@{}l@{}}A\_1: 86.8  \\ A\_2: 55.0  \\ A\_C: 23.6\end{tabular}} & \begin{tabular}[c]{@{}l@{}}A\_1: 90.2  \\ A\_2: 74.3  \\ A\_C: 45.5\end{tabular} & \begin{tabular}[c]{@{}l@{}}A\_1: 80.7  \\ A\_2: 9.8  \\ A\_C: 0.0\end{tabular}  & \begin{tabular}[c]{@{}l@{}}A\_1: 85.9  \\ A\_2: 40.0  \\ A\_C: NA\end{tabular} & \begin{tabular}[c]{@{}l@{}}A\_1: 81.6  \\ A\_2: 68.0  \\  A\_C: 68.1 \end{tabular}  \\ \hline
Llama-70B                                               & \begin{tabular}[c]{@{}l@{}}A\_1: 88.3  \\ A\_2: 16.4  \\ A\_C: 0.0\end{tabular} & \multicolumn{1}{l|}{\begin{tabular}[c]{@{}l@{}}A\_1: 93.6  \\ A\_2: 85.1  \\ A\_C: 76.4\end{tabular}} & \multicolumn{1}{l|}{\begin{tabular}[c]{@{}l@{}}\textbf{A\_1: 94.1}  \\ \textbf{A\_2: 88.3}  \\ \textbf{A\_C: 86.7}\end{tabular}} & \begin{tabular}[c]{@{}l@{}}A\_1: 0.1  \\ A\_2: 0.0  \\ A\_C: 0.0\end{tabular} & \multicolumn{1}{l|}{\begin{tabular}[c]{@{}l@{}}A\_1: 92.3  \\ A\_2: 66.7  \\ A\_C: 26.7\end{tabular}} & \begin{tabular}[c]{@{}l@{}}A\_1: 93.4  \\ A\_2: 83.6  \\ A\_C: 52.5\end{tabular}                                                                                                                                                    & \begin{tabular}[c]{@{}l@{}}A\_1: 75.8  \\ A\_2: 16.8  \\ A\_C: 0.0\end{tabular} & \begin{tabular}[c]{@{}l@{}}A\_1: 91.5  \\ A\_2: 45.8  \\ A\_C: NA\end{tabular} & \begin{tabular}[c]{@{}l@{}}A\_1: - \\ A\_2: -  \\  A\_C: - \end{tabular}  \\ \hline
MPT-7B                                                  & \begin{tabular}[c]{@{}l@{}}A\_1: 80.3  \\ A\_2: 2.7  \\ A\_C: 0.0\end{tabular}  & \multicolumn{1}{l|}{\begin{tabular}[c]{@{}l@{}}A\_1: 78.2  \\ A\_2: 42.3  \\ A\_C: 43.1\end{tabular}} & \multicolumn{1}{l|}{\begin{tabular}[c]{@{}l@{}}A\_1: 74.0  \\ A\_2: 49.3  \\ \textbf{A\_C: 54.1}\end{tabular}} & \begin{tabular}[c]{@{}l@{}}A\_1: 0.1  \\ A\_2: 0.0  \\ A\_C: 0.0\end{tabular} & \multicolumn{1}{l|}{\begin{tabular}[c]{@{}l@{}}A\_1: 75.7  \\ A\_2: 35.7  \\ A\_C: 13.9\end{tabular}} & \begin{tabular}[c]{@{}l@{}}A\_1: 70.9  \\ A\_2: 30.0  \\ A\_C: 10.0\end{tabular} & \begin{tabular}[c]{@{}l@{}}A\_1: 77.6  \\ A\_2: 4.8  \\ A\_C: 0.0\end{tabular}  & \begin{tabular}[c]{@{}l@{}}\textbf{A\_1: 82.7}  \\ \textbf{A\_2: 56.6}  \\ A\_C: NA\end{tabular} & \begin{tabular}[c]{@{}l@{}}A\_1: 61.0  \\ A\_2: 21.0  \\  A\_C: 9.6 \end{tabular}  \\ \hline
Falcon-7B                                               & \begin{tabular}[c]{@{}l@{}}A\_1: 63.2  \\ A\_2: 16.6  \\ A\_C: 0.0\end{tabular} & \multicolumn{1}{l|}{\begin{tabular}[c]{@{}l@{}}A\_1: 50.6  \\ A\_2: 35.6  \\ A\_C: 37.5\end{tabular}} & \multicolumn{1}{l|}{\begin{tabular}[c]{@{}l@{}}A\_1: 70.3  \\ \textbf{A\_2: 45.8}  \\ \textbf{A\_C: 53.0}\end{tabular}} & \begin{tabular}[c]{@{}l@{}}A\_1: 0.0  \\ A\_2: 0.0  \\ A\_C: 0.0\end{tabular} & \multicolumn{1}{l|}{\begin{tabular}[c]{@{}l@{}}A\_1: 54.7  \\ A\_2: 30.0  \\ A\_C: 27.1\end{tabular}} & \begin{tabular}[c]{@{}l@{}}A\_1: 69.9  \\ A\_2: 44.0  \\ A\_C: 42.0\end{tabular} & \begin{tabular}[c]{@{}l@{}}A\_1: 61.3  \\ A\_2: 10.8  \\ A\_C: 0.0\end{tabular} & \begin{tabular}[c]{@{}l@{}}\textbf{A\_1: 71.8}  \\ A\_2: 38.3  \\ A\_C: NA\end{tabular} & \begin{tabular}[c]{@{}l@{}}A\_1: 71.6  \\ A\_2: 45.6  \\  A\_C: 46.3 \end{tabular}  \\ \hline

\end{tabular}
\caption{DisentQA-DupliCite Results. 
The Document Split method involves
providing each document individually to the models, and hence, citations are known by default.}
% A\_1=Average accuracy when at least one of the two answers exist in the generated text; A\_2=Average accuracy when both of the answers exist in the generated text; A\_C=Average citation accuracy. Accuracies are shown in percentages. 
\label{dupli_disentqa_table}
\end{table*}

\begin{table}[t]
\centering
\small
\setlength{\tabcolsep}{2.0pt}
\begin{tabular}{|c|c|c|c|c|}
\hline
\begin{tabular}[c]{@{}c@{}}\textbf{Method} /\\ \textbf{Model}\end{tabular} & \textbf{Zero-Shot} & 
\begin{tabular}[c]{@{}c@{}}\textbf{C.A} \\ \textbf{Basic}\end{tabular} & \begin{tabular}[c]{@{}c@{}}\textbf{Few-shot} \\ \textbf{C.A CoT}\end{tabular} & \textbf{Finetuning} \\ \hline

Llama-7B & \begin{tabular}[c]{@{}c@{}}A\_1: 69.6  \\ A\_2: 7.3  \\ A\_C: 0.0\end{tabular} & \begin{tabular}[c]{@{}c@{}}A\_1: 74.3  \\ A\_2: \textbf{56.0} \\ A\_C: \textbf{59.0}\end{tabular} & \begin{tabular}[c]{@{}c@{}}A\_1: 65.3  \\ A\_2: 47.6  \\ A\_C: 35.0\end{tabular} & \begin{tabular}[c]{@{}c@{}}A\_1: \textbf{74.6}  \\ A\_2: 54.0  \\ A\_C: 40.0\end{tabular} \\ \hline

Llama-13B & \begin{tabular}[c]{@{}c@{}}A\_1: 71.6  \\ A\_2: 4.3  \\ A\_C: 0.0\end{tabular} & \begin{tabular}[c]{@{}c@{}}A\_1: 77.0  \\ A\_2: 58.0  \\ A\_C: 60.6\end{tabular} & \begin{tabular}[c]{@{}c@{}}A\_1: 72.0  \\ A\_2: 40.6  \\ A\_C: 10.8\end{tabular} & \begin{tabular}[c]{@{}c@{}}A\_1: \textbf{81.6}  \\ A\_2: \textbf{66.6}  \\ A\_C: \textbf{67.1}\end{tabular} \\ \hline

MPT-7B & \begin{tabular}[c]{@{}c@{}}A\_1: \textbf{65.3}  \\ A\_2: 0.3  \\ A\_C: 0.0\end{tabular} & \begin{tabular}[c]{@{}c@{}}A\_1: 54.3  \\ A\_2: \textbf{26.6}  \\ A\_C: 32.0\end{tabular} & \begin{tabular}[c]{@{}c@{}}A\_1: 56.0  \\ A\_2: 14.3  \\ A\_C: 4.3\end{tabular} & \begin{tabular}[c]{@{}c@{}}A\_1: 64.6  \\ A\_2: 22.6  \\ A\_C: \textbf{10.8}\end{tabular} \\ \hline

Falcon-7B & \begin{tabular}[c]{@{}c@{}}A\_1: 50.6  \\ A\_2: 7.6  \\ A\_C: 0.0\end{tabular} & \begin{tabular}[c]{@{}c@{}}A\_1: 54.3  \\ A\_2: 19.3  \\ A\_C: 30.3\end{tabular} & \begin{tabular}[c]{@{}c@{}}A\_1: 58.3  \\ A\_2: 22.0  \\ A\_C: 30.1\end{tabular} & \begin{tabular}[c]{@{}c@{}}A\_1: \textbf{69.3}  \\ A\_2: \textbf{41.6}  \\ A\_C: \textbf{40.1}\end{tabular} \\ \hline

\end{tabular}
\caption{DisentQA-ParaCite. 
% A\_1=Average accuracy when at least one of the two answers exist in the generated text; A\_2=Average accuracy when both of the answers exist in the generated text; A\_C=Average citation accuracy. Accuracies are shown in percentages. 
C.A=Conflict-aware. We use 3 examples for both C.A Basic and CoT.}
\label{para_disentqa_table}
\end{table}

We next analyze the ability of models to answer questions with ambiguous contexts on the DisentQA-DupliCite and DisentQA-ParaCite datasets. Results can be seen in Tables \ref{dupli_disentqa_table} and \ref{para_disentqa_table}.

Out-of-the-box models are unable to generate citations, and generally struggle to produce multiple answers, resulting in poor A\_2 scores. Most methods improve models' A\_1 scores and their ability to generate distinct responses, with the best prompting method being C.A basic using 3 in-context examples. In contrast, the Zero-shot CoT method performs poorly. We also find that with 5 examples, the performance on DisentQA-DupliCite drops due to context size exceeding the models' maximum capacity, leading to test question truncation.

Notably, models' scores are significantly higher on the DisentQA-DupliCite dataset, with A\_1 scores ranging from $70.3\%$ to $94.1\%$ using the C.A basic method (3-shot), compared to $39.7\%$ to $76.2\%$ on AmbigQA-Cite. The document split method improves all models' performances, but only outperforms the few-shot method for MPT-7B and Falcon-7B models on A\_1.

In contrast, the DisentQA-ParaCite dataset presents a more challenging scenario, with overall lower scores than on DisentQA-DupliCite. However, we observe similar behavior, with C.A basic and finetuning methods yielding comparable scores. Interestingly, finetuning emerges as the overall best method on DisentQA-ParaCite.

\subsection{Ambiguous Context: Multi-hop}
\label{Ambiguous_Context_Multi}
\begin{table}[t]
\centering
\small
\setlength{\tabcolsep}{2.0pt}
\begin{tabular}{|c|l|l|l|l|}
\hline
\begin{tabular}[c]{@{}c@{}}\textbf{Method}/\\ \textbf{Model}\end{tabular} & \multicolumn{1}{c|}{\textbf{Zero-Shot}}                          & \multicolumn{1}{c|}{\textbf{\begin{tabular}[c]{@{}c@{}}C.A\\ Basic\end{tabular}}}             
& \multicolumn{1}{c|}{\textbf{\begin{tabular}[c]{@{}c@{}}Few-shot\\ C.A CoT\end{tabular}}}
& \multicolumn{1}{c|}{\textbf{Finetuning}}
\\ \hline
Llama-7B                                                & \begin{tabular}[c]{@{}l@{}}A\_1: 82.6  \\ A\_2: 27.6  \\ A\_3: 5.0 \\ A\_C: 0.0\end{tabular} 
& \begin{tabular}[c]{@{}l@{}}A\_1: 67.0  \\ A\_2: 36.1  \\ A\_3: 10.3 \\ A\_C: 8.9\end{tabular} 
& \begin{tabular}[c]{@{}l@{}}A\_1: 75.0  \\ A\_2: 25.0  \\ A\_3: 10.0 \\ A\_C: 11.6\end{tabular} 
& \begin{tabular}[c]{@{}l@{}}A\_1: \textbf{98.0}  \\ A\_2: \textbf{90.0}  \\ A\_3: \textbf{62.0 }\\ A\_C: \textbf{67.3} \end{tabular}  \\ \hline
Llama-13B                                               & \begin{tabular}[c]{@{}l@{}}A\_1: 83.0  \\ A\_2: 21.3  \\ A\_3: 4.6 \\ A\_C: 0.0\end{tabular} 
& \begin{tabular}[c]{@{}l@{}}A\_1: 86.0  \\ A\_2: 68.9  \\ A\_3: 37.0 \\ A\_C: 36.3\end{tabular} 
& \begin{tabular}[c]{@{}l@{}}A\_1: 80.0  \\ A\_2: 55.0  \\ A\_3: 25.0 \\ A\_C: 13.3\end{tabular} 
& \begin{tabular}[c]{@{}l@{}}A\_1: \textbf{98.3}  \\ A\_2: \textbf{93.3}  \\ A\_3: \textbf{65.6} \\ A\_C: \textbf{76.3} \end{tabular}  \\ \hline
MPT-7B                                                  & \begin{tabular}[c]{@{}l@{}}A\_1: 72.3  \\ A\_2: 16.0  \\ A\_3: 2.0 \\ A\_C: 0.0\end{tabular} 
& \begin{tabular}[c]{@{}l@{}}A\_1: 65.3  \\ A\_2: 24.0  \\ A\_3: 4.8 \\ A\_C: 0.03\end{tabular} 
& \begin{tabular}[c]{@{}l@{}}A\_1: 50.0  \\ A\_2: 15.0  \\ A\_3: 0.0 \\ A\_C: 0.0\end{tabular} 
& \begin{tabular}[c]{@{}l@{}}A\_1: \textbf{93.0}  \\ A\_2: \textbf{84.3}  \\ A\_3: \textbf{59.0} \\ A\_C: \textbf{64.5} \end{tabular}  \\ \hline
Falcon-7B                                               & \begin{tabular}[c]{@{}l@{}}A\_1: 63.0  \\ A\_2: 24.6  \\ A\_3: 6.3 \\ A\_C: 0.0\end{tabular}  
& \begin{tabular}[c]{@{}l@{}}A\_1: 48.1  \\ A\_2: 15.4  \\ A\_3: 2.6 \\ A\_C: 0.01\end{tabular} 
& \begin{tabular}[c]{@{}l@{}}A\_1: 0.0  \\ A\_2: 0.0  \\ A\_3: 0.0 \\ A\_C: 0.0\end{tabular} 
& \begin{tabular}[c]{@{}l@{}}A\_1: \textbf{85.3}  \\ A\_2: \textbf{75.3 } \\ A\_3: \textbf{39.3 }\\ A\_C: \textbf{49.7} \end{tabular}  \\ \hline
\end{tabular}
\caption{Conflicting HotpotQA-Cite (no distractors). C.A=Conflict-aware. We use 3 examples for both C.A Basic and CoT. 
% Zero-shot vs. the best prompting method -- conflict-aware prompting, using 3 in-context examples. A\_1, A\_2, and A\_3 are the average accuracy when at least one, two, or three of the answers exist in the generated text, respectively; A\_C=Average citation accuracy. Accuracies are shown in percentages.
}
\label{no_distractors_hotpotqa_table}
\end{table}
\begin{table}[t]
\centering
\small
\setlength{\tabcolsep}{2.0pt}
\begin{tabular}{|c|l|l|l|l|}
\hline
\begin{tabular}[c]{@{}c@{}}\textbf{Method}/\\ \textbf{Model}\end{tabular} & \multicolumn{1}{c|}{\textbf{Zero-Shot}}                          & \multicolumn{1}{c|}{\textbf{\begin{tabular}[c]{@{}c@{}}C.A\\ Basic\end{tabular}}}             
& \multicolumn{1}{c|}{\textbf{\begin{tabular}[c]{@{}c@{}}Few-shot\\ C.A CoT\end{tabular}}}
& \multicolumn{1}{c|}{\textbf{Finetuning}}
\\ \hline
Llama-7B                                                & \begin{tabular}[c]{@{}l@{}}A\_1: \textbf{59.8}  \\ A\_2: 16.8  \\ A\_3: 2.2 \\ A\_C: 0.0\end{tabular} 
& \begin{tabular}[c]{@{}l@{}}A\_1: 38.0  \\ A\_2: 9.3  \\ A\_3: 1.0 \\ A\_C: 0.1\end{tabular} 
& \begin{tabular}[c]{@{}l@{}}A\_1: 39.3  \\ A\_2: 12.6  \\ A\_3: 0.6 \\ A\_C: 0.2\end{tabular} 
& \begin{tabular}[c]{@{}l@{}}A\_1: 49.0  \\ A\_2: \textbf{17.3}  \\ A\_3: \textbf{2.3}\\ A\_C: \textbf{2.0} \end{tabular}  \\ \hline
Llama-13B                                               & \begin{tabular}[c]{@{}l@{}}A\_1:\textbf{ 60.8}  \\ A\_2: 15.8  \\ A\_3: 3.5 \\ A\_C: 0.0\end{tabular} 
& \begin{tabular}[c]{@{}l@{}}A\_1: 51.0  \\ A\_2: \textbf{20.0}  \\ A\_3: \textbf{2.6} \\ A\_C: \textbf{1.3}\end{tabular} 
& \begin{tabular}[c]{@{}l@{}}A\_1: 42.3  \\ A\_2: 13.0  \\ A\_3: 2.3 \\ A\_C: 0.1\end{tabular} 
& \begin{tabular}[c]{@{}l@{}}A\_1: 46.6  \\ A\_2: 16.3  \\ A\_3: 2.0 \\ A\_C: 1.2 \end{tabular}  \\ \hline
MPT-7B                                                  & \begin{tabular}[c]{@{}l@{}}A\_1: \textbf{49.5}  \\ A\_2: 13.0  \\ A\_3: \textbf{3.0 }\\ A\_C: 0.0\end{tabular} 
& \begin{tabular}[c]{@{}l@{}}A\_1: 31.0  \\ A\_2: 8.6  \\ A\_3: 1.0 \\ A\_C: 0.0\end{tabular} 
& \begin{tabular}[c]{@{}l@{}}A\_1: 42.0  \\ A\_2: 12.3  \\ A\_3: 1.3 \\ A\_C: 0.0\end{tabular} 
& \begin{tabular}[c]{@{}l@{}}A\_1: 48.6  \\ A\_2: \textbf{15.3}  \\ A\_3: 1.6 \\ A\_C: \textbf{0.4 }\end{tabular}  \\ \hline
Falcon-7B                                               & \begin{tabular}[c]{@{}l@{}}A\_1: 29.5  \\ A\_2: 7.5  \\ A\_3: 2.5 \\ A\_C: 0.0\end{tabular}  
& \begin{tabular}[c]{@{}l@{}}A\_1: 25.3  \\ A\_2: 8.0  \\ A\_3: 1.3 \\ A\_C: 0.0\end{tabular} 
& \begin{tabular}[c]{@{}l@{}}A\_1: 4.3  \\ A\_2: 0.0  \\ A\_3: 0.0 \\ A\_C: 0.0\end{tabular} 
& \begin{tabular}[c]{@{}l@{}}A\_1: \textbf{37.4}  \\ A\_2:\textbf{ 8.4} \\ A\_3: \textbf{8.6}\\ A\_C: \textbf{0.2} \end{tabular}  \\ \hline
\end{tabular}
\caption{Conflicting HotpotQA-Cite (Distractors). C.A=Conflict-aware. We use 3 examples for both C.A Basic and CoT.}
\label{distractos_hotpotqa_table}
\end{table}
% We do not evaluate the document split method here, as the entire premise of multi-hop reasoning is that multiple documents are required to answer the question, which the document split method cannot handle as by design only one document is visible to models at a time. Additionally, while it is theoretically possible to use more than one document in this method, it is intractable to get every permutation of all documents, as the number of possible answers is unknown, and the amount of documents may be large. 
% For example, all models' A\_1, A\_2, and A\_3 scores greatly improves, with the largest increase on A\_1 occurs for Llama-13B ($60.8\%$ to $86.0\%$), the largest on A\_2 occurs for Llama-70B ($20.0\%$ to $76.0\%$), and the largest on A\_3 occurs for Llama-70B ($4.5\%$ to $46.3\%$). We also find a large increase in models' A\_C score, however, only for the Llama models. For example, while the increase for Llama-7B, 13B, and 70B are $8.9\%$, $36.3\%$, and $45.2\%$, respectively, the increase for MPT-7B and Falcon-7B are only $0.03\%$, and $0.01\%$. 

We evaluate our baselines on the more complex Conflicting HotpotQA-Cite dataset, which involves multi-hop QA with many conflicting answers. The results are presented in Tables \ref{no_distractors_hotpotqa_table} and \ref{distractos_hotpotqa_table}.

On the no-distractor variant dataset, we observe two unexpected trends. While the C.A basic method improves models' performances on A\_2, A\_3, and A\_C metrics, it underperforms the zero-shot baseline on A\_1. In contrast, finetuning significantly outperforms all other methods, achieving nearly $100\%$ A\_1 scores across Llama-7B, Llama-13B, and MPT-7B. However, all models are still far from perfect on generating all correct answers correctly, in addition to citing their sources. Lastly, the few-shot CoT method generally performs poorly across all models and metrics.

On the distractor variant dataset, the C.A basic method underperforms the zero-shot baseline on A\_1, but outperforms it on A\_2, A\_3, and A\_C. The overall models' scores are significantly lower than on the no-distractor setting, indicating that this setting is more challenging for models. Finetuning again emerges as the best approach, outperforming most methods. However, all baselines struggle to generate multiple correct answers, with the best scores being $17.3\%$ for A\_2 (finetuned Llama-7B) and $8.6\%$ for A\_3 (finetuned Falcon-7B). Additionally, they perform poorly on citing their sources.

\subsection{Non-ambiguous Context: Single-hop}
\begin{table}[t]
\centering
\small
\setlength{\tabcolsep}{2.0pt}
\begin{tabular}{|c|l|l|l|}
\hline
\textbf{\begin{tabular}[c]{@{}c@{}}Method/\\ Model\end{tabular}} & \multicolumn{1}{c|}{\textbf{Zero-Shot}} & \multicolumn{1}{c|}{\textbf{\begin{tabular}[c]{@{}c@{}}C.A\\ Basic\end{tabular}}} & \multicolumn{1}{c|}{\textbf{Finetuning}} \\ \hline
Llama-7B                                                         & \textbf{A\_1: 84.8}                             & A\_1: 84.1                                                                               & A\_1: 64.3   \\ \hline
Llama-13B                                                        & \textbf{A\_1: 83.8}                             & A\_1: 80.3                                                                               & A\_1: 71.6  \\ \hline
Llama-70B                                                        & \textbf{A\_1: 89.0}                             & A\_1: 88.1                                                                               & A\_1: -  \\ \hline
MPT-7B                                                           & \textbf{A\_1: 81.2}                             & A\_1: 73.9                                                                               & A\_1: 51.3  \\ \hline
Falcon-7B                                                        & A\_1: 71.1                             & \textbf{A\_1: 72.5}                                                                               & A\_1: 46.3  \\ \hline
\end{tabular}
\caption{DisentQA with no contextual conflicts. C.A=Conflict-aware. We use 3 examples for C.A Basic.
% Zero-shot vs. the best prompting method -- conflict-aware prompting, using 3 in-context examples. A\_1=Average accuracy when the correct answer exists in the generated text.
}
\label{no_conflict_table}
\end{table}
\label{Non-ambiguous_Context}
We assess whether the top-performing techniques, C.A basic and finetuning, degrade models' performances compared to the zero-shot baseline when no ambiguity exists. We use the original context from the DisentQA dataset, which lacks knowledge conflicts. The results are presented in Table \ref{no_conflict_table}.

For the C.A basic method, we observe that most models experience some performance degradation, except for Falcon-7B, which actually shows a performance increase. For example, MPT-7B suffers the largest A\_1 drop, from $81.2\%$ to $73.9\%$, while Llama-7B experiences the smallest drop, from $84.8\%$ to $84.1\%$. However, this performance drop is relatively small compared to the significant gains provided by this method in Sections \ref{Ambiguous_questions}, \ref{Ambiguous_Context_Single}, and \ref{Ambiguous_Context_Multi}. In contrast, finetuning results in a much more substantial performance drop. For instance, Falcon-7B's A\_1 score plummet from $71.1\%$ to $46.3\%$.

\section{Discussion}
\subsection{Ambiguous Questions vs. Contexts}
\label{ambigqa_vs_disentqa}
% On AmbigQA-Cite, the top scores for (A\_1), (A\_2), and (A\_C) are $76.3%$, $35.4%$, and $48.3%$, respectively (see Table \ref{ambigqa_table}). In contrast, the top scores on DisentQA are (A_1), (A_2), and (A_C) are $94.1%$, $88.3%$, and $86.7%$ (see Table \ref{disentqa_table}).

We observe a significant performance gap between the AmbigQA-Cite and DisentQA-DupliCite datasets. This disparity can be attributed to two primary factors. 1) DisentQA-DupliCite is constructed using the entity-substitution method, which generates two contexts with a single differing entity answer. This design makes the task relatively easier compared to AmbigQA-Cite, where no duplicates exist. 2) AmbigQA-Cite's questions are intentionally ambiguous, rendering them more challenging to answer than those in DisentQA-DupliCite. Moreover, we observe that models perform worse on DisentQA-ParaCite, suggesting that paraphrased contexts introduce a higher level of complexity compared to entity substitution, which helps to bridge the performance gap.

\subsection{Multi-hop vs. Single Hop}
% Although the DisentQA-DupliCite and conflicting HotpotQA datasets differ in several ways, they share a similar approach to creating conflicting contexts: replacing the answer string with a different string, resulting in contexts with duplicated content. 
DisentQA-DupliCite and conflicting HotpotQA datasets share a common approach to creating conflicting contexts: replacing the answer string with a different string, yielding duplicated content. Comparing the results in Tables \ref{dupli_disentqa_table} and \ref{no_distractors_hotpotqa_table}, we observe two significant trends: firstly, generating correct citations is much more challenging in the multi-hop setting, where multiple documents exist and are required to reach the answer. Secondly, producing all correct answers is also much harder, even with a limited number of correct ones. Moreover, the presence of distractors in the conflicting HotpotQA dataset further exacerbates this challenge, leading to an even more significant performance drop. These results underscore the importance of developing novel conflicting multi-hop QA datasets.

% While the DisentQA-DupliCite and conflicting HotpotQA datasets are different in many ways, the conflicting contexts in both are created using a similar approach: by replacing the answer string with a different string (either using entity-substitution or MLM), and hence, each question corresponds to a long context which is partially composed  of context-duplication.

% Comparing Table \ref{disentqa_table} and Table \ref{no_distractors_hotpotqa_table}, we find two notable things: 1) correct citations are much harder to generate in the multi-hop setting, where both more documents exist and also multiple documents are required to reach the answer. This can be seen by observing that the lowest and highest scores on $A\_C$ on DisentQA are $53.0\%$ and $86.7\%$, respectively, where on the conflicting HotpotQA they are $0.01\%$ and $42.2\%$. 2) Producing all of the correct answers is also much harder, even with only 3 correct ones. For example, the best performing model on the conflicting HotpotQA dataset -- Llama-70B -- has a performance drop of $29.7\%$ going from A\_2 to A\_3 ($76\%$ to $46.3\%$). When added distractors to the conflicting HotpotQA the drop is even more significant. These results further highlight why developing such novel conflicting multi-hop QA datasets is important.

\subsection{3-shot vs. 5-shot}
While on the AmbigQA dataset we see a drop in performance across all models between the 3-shot and 5-shot few-shot method (see Table \ref{ambigqa_table}), the performance drop is far more significant in Table \ref{dupli_disentqa_table} on the DisentQA dataset. Analyzing this further, we find that with 5 examples the context becomes larger (especially on the DisentQA dataset) than the maximum context length the models can handle, which results in the test question truncation.

% While we observe a performance drop across all models from 3-shot to 5-shot few-shot method on AmbigQA-Cite (Table \ref{ambigqa_table}), the drop is more pronounced on DisentQA (Table \ref{dupli_disentqa_table}). This is because with 5 examples, the context exceeds the models' maximum context size, leading to test question truncation, especially on DisentQA.

\subsection{C.A Prompting vs. Zero-shot CoT}
One possible reason for the zero-shot CoT's poor performance on A\_C, with a score of $0\%$ across all models and tested datasets, is that it lacks an explicit citation prompt. Unlike the C.A methods, which specifically ask models to cite their sources, the zero-shot method only generates a reasoning chain in the first step, without explicitly requesting citation. This highlights the necessity of a specific citation prompt. Furthermore, we observe a significant difference in A\_2 scores between the two methods in both Tables \ref{ambigqa_table} and \ref{dupli_disentqa_table}, suggesting that models' self-generated reasoning chains are insufficient to handle conflicting facts.

\subsection{Limited Efficacy of C.A. Prompts on HotpotQA}
We find that both the C.A basic and C.A CoT perform worse than the zero-shot baseline and finetuning approach on the conflicting HotpotQA-Cite datasets. We hypothesize that this may be due to several reasons, such as the complexity of the multi-hop contexts, more cited documents in the multi-hop dataset, or that the in-context examples in the multi-hop setting were not as beneficial. 

\subsection{Finetuning vs. Prompting}
% Previous work has shown that prompting methods can improve model performance without finetuning \cite{NEURIPS2020_1457c0d6, liu2021pretrain, wei2023chainofthought, dong2023survey}. Our results reveal that the C.A prompting method often performs comparably to finetuning on AmbigQA-Cite and DisentQA-ParaCite, and even outperforms it on DisentQA-DupliCite. However, it struggles on Conflicting HotpotQA-Cite. Notably, finetuned models degrade significantly when no conflicts exist. Overall, we conclude that the C.A basic method is the best approach, however, both methods can be improved (see Section \ref{future_work}).

Consistent with prior work \cite{NEURIPS2020_1457c0d6, liu2021pretrain, wei2023chainofthought, dong2023survey}, our results show that the C.A prompting method can achieve comparable or even better performance than finetuning on AmbigQA-Cite, DisentQA-ParaCite, and DisentQA-DupliCite. However, it struggles on Conflicting HotpotQA-Cite. Notably, finetuned models experience significant degradation when no conflicts exist. Overall, we conclude that the C.A basic method is the most effective approach, but both methods have room for improvement (see Section \ref{future_work}).

% Many previous work have shown that prompting methods have great success in improving models' performance without the need for further finetuning \cite{NEURIPS2020_1457c0d6, liu2021pretrain, wei2023chainofthought, dong2023survey}. Here, we find that the C.A prompting method often scores comparable to the finetuning approach such as on the AmbigQA-Cite (Table \ref{ambigqa_table}) and DisentQA-ParaCite (Table \ref{para_disentqa_table})datasets. We also see that the C.A method outperform the finetunining approach on the DisentQA-DupliCite dataset (Table \ref{dupli_disentqa_table}), but perform much worse on the Conflicting HotpotQA-Cite (no distractor) dataset (Table \ref{no_distractors_hotpotqa_table}). Notably, it is important to see that when no conflict exist the performance of the finetuned models degrades significantly (Table \ref{no_conflict_table}). To that end, we believe the overall best method is the C.A basic.

\section{Real-world Usage}
In our comprehensive analysis, we evaluate three main approaches to improve LLMs' ability to answer ambiguous questions with source citations: 1) prompt-based; 2) rule-based, and 3) fine-tuning-based. Notably, while the rule-based approach outperforms the other two in some occasions, as discussed in Section \ref{doc_split}, it is incapable of answering questions that require complex reasoning across multiple documents, as it only sees one document at a time. To that end, we do not recommend using this approach when it is known that the data is of complex nature. But, to use it, users need to split the retrieved documents into chunks of one document at a time, which are sent to the model, followed by an aggregation of the answers. With regards to the other two approaches, the prompt-based approaches can be incorporated into most LLMs with a simple addition of a prompt, as shown in Appendix \ref{prompt_design}. However, it is worth mentioning that the fine-tuning approach outperforms the prompting approach on multihop reasoning, but also results in a large performance decrease when no ambiguity exists, as discussed in Section \ref{Non-ambiguous_Context}. We also showed that LoRA-based fine-tuning is sufficient to improve LLMs' abilities in this task greatly over the baseline, highlighting the usability for real-users that do not have large computational resources.

\section{Future Work}
\label{future_work}
% finetuning with distractors -- we only trained hotpot on the no-distractor setting
% finetuning with both conflicting and non-conflicting data -- we only trained on conflicting data
% finetuning on paraphrase instead of duplications
% other prompts that emphasize citations or other citation methods
% creating more + different datasets -- we only experiment with 3 flavors: paraphrasing, distractors, and duplications
% different architecture, larger models 
% pretraining on conflicting data and citations

Having established the novel task of QA with source citation in ambiguous settings, and proposing strong baselines across our newly created datasets, several promising directions for future research emerge. Future work can build upon our contributions by: 1) Investigating the impact of finetuning on datasets with distractors, as our current experiments only focused on the no-distractor setting; 2) extending our finetuning data to include both conflicting and non-conflicting instances, which could lead to more robust models capable of handling varying levels of ambiguity; 3) exploring the effectiveness of finetuning on paraphrased data instead of duplications, which may provide additional insights into the model's ability to generalize across different linguistic formulations; 4) designing alternative prompts that emphasize the importance of citations or exploring other citation methods to further enhance the model's performance; 5) creating a more diverse range of datasets that capture different aspects of ambiguity, beyond the three flavors (paraphrasing, distractors, and duplications) we experiment with in this work; 6) investigating the potential benefits of employing different architectures or larger language models to tackle this challenging task; and 7) pretraining models on conflicting data and citations to potentially improve their ability to resolve knowledge conflicts and provide trustworthy answers.

% By pursuing these research directions, we hope to further advance the state-of-the-art in QA and develop more reliable, interpretable, and comprehensive systems that can effectively navigate ambiguous settings.

\section{Conclusion}
% We address a significant gap in QA research by introducing the novel task of QA with source citation in ambiguous settings. This task combines the complexities of ambiguous answers with the importance of providing source citations, enabling users to evaluate the factuality of each answer. To facilitate research in this area, we provide a comprehensive framework consisting of novel datasets, new evaluation metrics, and strong baselines using various approaches over 5 LLMs. Our work aims to inspire the development of more trustworthy and interpretable QA systems, bridging the gap between ambiguous answer resolution and citation generation.

We address a significant gap in QA research by introducing the novel task of QA with source citation in ambiguous settings. This task combines the complexities of ambiguous QA with the importance of providing source citations, enabling users to evaluate the factuality of each answer. To facilitate research in this area, we provide a comprehensive framework consisting of novel datasets, new evaluation metrics, and strong baselines using various approaches over 5 LLMs. Our work aims to inspire the development of more trustworthy and interpretable QA systems, bridging the gap between ambiguous answer resolution and citation generation. By exploring this new task, we hope to pave the way for more reliable and transparent QA systems that can accurately resolve knowledge conflicts and provide users with credible sources to support their answers.

\section*{Limitations}
While we have evaluated 5 diverse LLMs and observed similar limitations, it is an open question whether other LLMs would face similar challenges in generating citations and handling multiple answers or ambiguous context. However, our findings suggest that many models may struggle with these tasks. Furthermore, our focus on unstructured texts from standard reading comprehension datasets raises the question of whether other knowledge formats, such as knowledge graph triples, would yield similar results.

\section*{Ethics Statement}
Our motivation is to create systems that can effectively handle conflicting information and provide source citations, enabling users to verify the accuracy of the answers. We emphasize the importance of future research building upon this foundation to develop reliable systems that can be safely deployed in real-world applications.

\bibliography{anthology,custom}
\bibliographystyle{acl_natbib}

\clearpage 

\appendix
% \section{Datasets Examples}
\section{Prompts Used}
\label{prompt_design}
\subsection{Conflict-aware Basic Prompting}
We use the following prompt design:

\noindent
\texttt{
You will get a question and some context texts. These texts may conflict with each other. If they do, answer according to the following examples: Example 1: Question: $q_{e1}$ Context: $c_{e1}$ Answer: According to Document D\_1 the answer is A\_1. According to Document D\_2 the answer is A\_2 [...] Question: $q_{ti}$ Context: $c_{ti}$. Answer:
}

\subsection{Few-shot Conflict-aware CoT Prompting}
\noindent
\texttt{
Example 1: Question: $q_{e1}$. Context: $c_{e1}$. Reasoning: Document D\_1 mentions that the answer is A\_1. But, Document D\_2 mentions that the answer is A\_2. Therefore, the answer is: According to Document D\_1 the answer is A\_1. According to Document D\_2 the answer is A\_2. Question: $q_{ti}$ Context: $c_{ti}$. Answer:
}

\subsection{Zero-shot CoT Prompting}

\noindent

\textbf{Step 1}
\begin{center}
\texttt{
Question: $q_{ti}$ Context: $c_{ti}$. Answer: Let’s think step by step.}
\end{center}

This produces a generated response $r_i$.

\textbf{Step 2}
\begin{center}
\texttt{
Question: $q_{t1}$ Context: $c_{t1}$. $r_i$. Therefore, the answer is}
\end{center}

\section{Finetuning Parameters}
\label{finetuning_parap}
Our fine-tuning settings include: LoRA with alpha=15 and dropout=0.1; 1000 steps with early stopping (patience=3) guided by the validation set; batch size=4; gradient accumulation steps=1; AdamW optimizer with a linear learning rate scheduler and 10 warmup steps.

% \subsection{AmbigQA}

% \subsection{DisentQA}

% \subsection{Conflicting HotpotQA}

\end{document}